\documentclass[10pt,twocolumn,letterpaper]{article}

\usepackage{iccv}
\usepackage{times}
\usepackage{epsfig}
\usepackage{graphicx}
\usepackage{amsmath}
\usepackage{amssymb}

\usepackage[table]{xcolor}

\usepackage{algorithm}
\usepackage{bm}
\usepackage{algorithmic}
\usepackage{listings}
\usepackage{minipage-marginpar}
\usepackage{etoolbox}
\makeatletter

\AfterEndEnvironment{algorithm}{\let\@algcomment\relax}
\AtEndEnvironment{algorithm}{\kern2pt\hrule\relax\vskip3pt\@algcomment}
\let\@algcomment\relax
\newcommand\algcomment[1]{\def\@algcomment{\footnotesize#1}}
\renewcommand\fs@ruled{\def\@fs@cfont{\bfseries}\let\@fs@capt\floatc@ruled
  \def\@fs@pre{\hrule height.8pt depth0pt \kern2pt}%
  \def\@fs@post{}%
  \def\@fs@mid{\kern2pt\hrule\kern2pt}%
  \let\@fs@iftopcapt\iftrue}
\makeatother

\usepackage[pagebackref=true,breaklinks=true,letterpaper=true,colorlinks,bookmarks=false]{hyperref}

 \iccvfinalcopy 

\ificcvfinal

\begin{document}

\title{DiffPose: SpatioTemporal Diffusion Model for Video-Based \\ Human Pose Estimation \vspace{-1.2em}}

\author{
    Runyang Feng\textsuperscript{\rm 1,2},
    Yixing Gao\textsuperscript{\rm 1,2}\thanks{Corresponding Author},
    Tze Ho Elden Tse\textsuperscript{\rm 3},
    Xueqing Ma\textsuperscript{\rm 1,2},
    Hyung Jin Chang\textsuperscript{\rm 3}
    \\    
    \textsuperscript{1} School of Artificial Intelligence, Jilin University,\\
    \textsuperscript{2} Engineering Research Center of Knowledge-Driven Human-Machine Intelligence, \\Ministry of Education, China,
    \textsuperscript{3}School of Computer Science, University of Birmingham
	\\
	{\tt\small \{fengry22, maxq21\}@mails.jlu.edu.cn,
	 \tt\small gaoyixing@jlu.edu.cn,}\\
	 {\tt\small txt994@student.bham.ac.uk, h.j.chang@bham.ac.uk
	}
}

\maketitle
\ificcvfinal

\begin{abstract} \vspace{-.5em}
    Denoising diffusion probabilistic models that were initially proposed for realistic image generation have recently shown success in various perception tasks (e.g., object detection and image segmentation) and are increasingly gaining attention in computer vision. However, extending such models to multi-frame human pose estimation is non-trivial due to the presence of the additional temporal dimension in videos.
    More importantly, learning representations that focus on keypoint regions is crucial for accurate localization of human joints. Nevertheless, the adaptation of the diffusion-based methods remains unclear on how to achieve such objective.
    In this paper, we present DiffPose, a novel diffusion architecture that formulates video-based human pose estimation as a conditional heatmap generation problem. 
    First, to better leverage temporal information, we propose SpatioTemporal Representation Learner which aggregates visual evidences across frames and uses the resulting features in each denoising step as a condition.
    In addition, we present a mechanism called Lookup-based MultiScale Feature Interaction that determines the correlations between local joints and global contexts across multiple scales. This mechanism generates delicate representations that focus on keypoint regions.
    Altogether, by extending diffusion models, we show two unique characteristics from DiffPose on pose estimation task: (i) the ability to combine multiple sets of pose estimates to improve prediction accuracy, particularly for challenging joints, and (ii) the ability to adjust the number of iterative steps for feature refinement without retraining the model.
    DiffPose sets new state-of-the-art results on three benchmarks: PoseTrack2017, PoseTrack2018, and PoseTrack21. 
\end{abstract}

\begin{figure}
\begin{center}
\includegraphics[width=.98\linewidth]{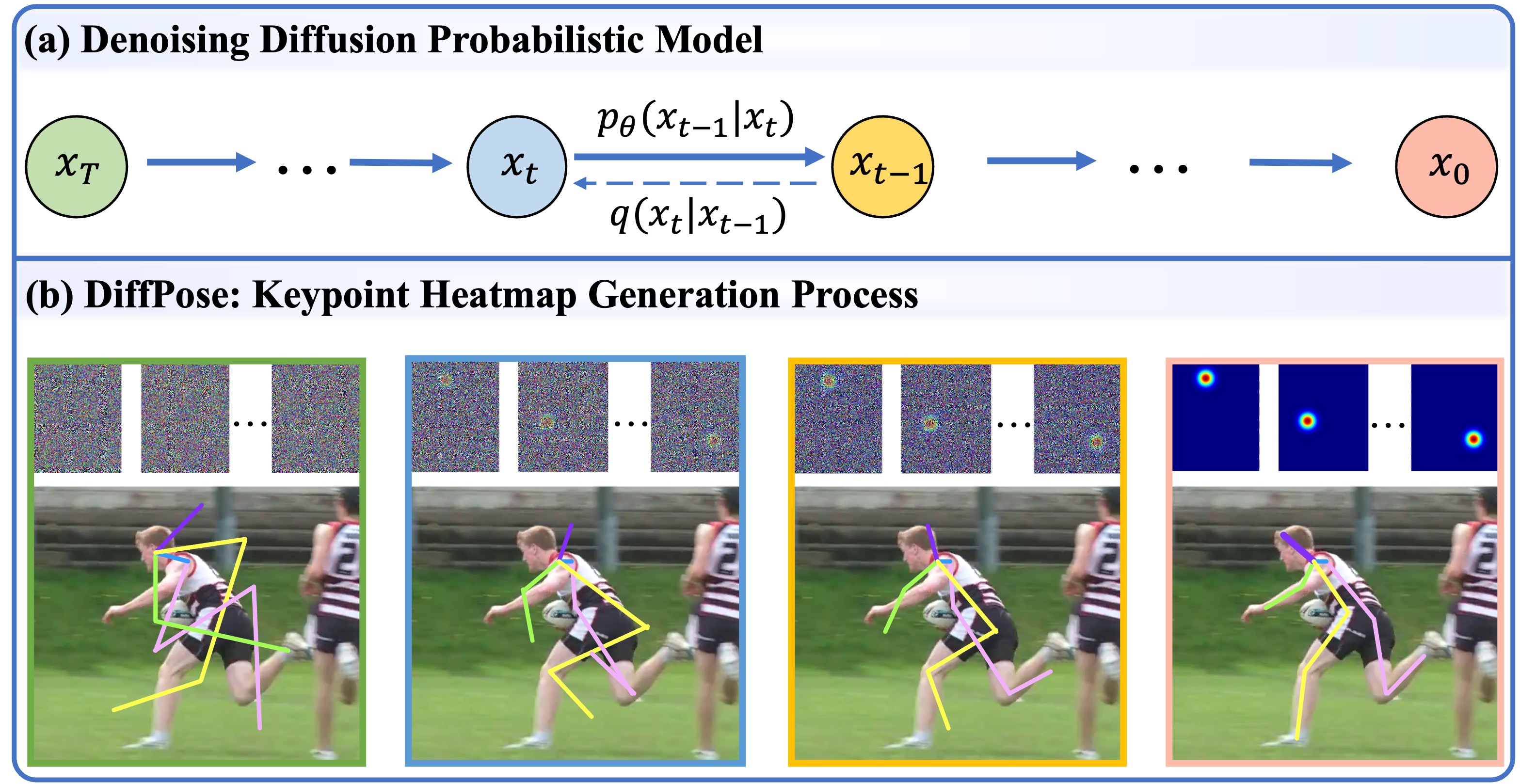}
\end{center}
\vspace{-1em}
\caption{
\textbf{(a)} Illustration of the original diffusion model where $q$ and $p_\theta$ refer to the diffusion and denoising process, respectively.
\textbf{(b)} In this work, we propose a novel framework named DiffPose which formulates video-based human pose estimation as a generative process of keypoint heatmaps.
}\vspace{-1.2em}
\label{fig:formulation}
\end{figure}

\vspace{-1.5em}
\section{Introduction}
Human pose estimation has been extensively studied in computer vision, with the aim of detecting all instances of people from images and localizing anatomical keypoints  for each individual \cite{feng2023mutual, shi2022end, sun2019deep, wang2022contextual}.
It finds numerous applications ranging from human-computer interaction and augmented reality to behavior analysis and surveillance tracking \cite{li2021human, liu2022temporal, schmidtke2021unsupervised, tse2019no, tse2022collaborative, tse2022s}.  
Conventional approaches \cite{sapp2010cascaded, wang2013beyond, zhang2009efficient} mainly employ the probabilistic graphical model or the pictorial structure model. 
Fueled by the explosion of deep learning, \emph{Convolutional Neural Networks} \cite{bertasius2019learning, liu2021deep, liu2022temporal, sun2019deep} and \emph{Vision Transformers} \cite{li2021tokenpose, xu2022vitpose, yuan2021hrformer} have witnessed significant progress in this task.

Until recently, denoising diffusion probabilistic models \cite{ho2020denoising, sohl2015deep}, which are a type of generative models, have received much research attention for surpassing other methods such as GANs and achieving state-of-the-art generative results \cite{baranchuk2021label, dhariwal2021diffusion}.
The superior performance of the diffusion model has facilitated its expansion in diverse applications, such as super-resolution \cite{saharia2022image}, inpainting \cite{lugmayr2022repaint}, and image deblurring \cite{ren2022image}.
Following the demonstration of the effectiveness of diffusion models as representation learners for discriminative computer vision problems \cite{baranchuk2021label}, several contemporary approaches have successfully employed the diffusion model for perception tasks, including object detection \cite{chen2022diffusiondet} and image segmentation \cite{amit2021segdiff, baranchuk2021label, gu2022diffusioninst}. 


Despite the considerable attention that diffusion models have gained following their achievements, their adaptation for video-based human pose estimation has significantly trailed that of other vision tasks, such as segmentation and object detection.
We conjecture two primary reasons that underlie this disparity:
 \textbf{(i)} Effectively leveraging temporal information is crucial for video-based human pose estimation \cite{liu2021deep}. 
 However, despite the success of various diffusion architectures in perception tasks, they are primarily designed for static images and incapable to capture temporal dependencies across frames.
\textbf{(ii)} Real world images typically contain many task-irrelevant cues and accurately estimating human poses requires focusing on specific body joint regions \cite{geng2021bottom}. However, it is still an open question on how to guide the diffusion model to filter out the unnecessary details and only attend to the keypoint regions.

In this paper, we present a novel architecture, termed SpatioTemporal \underline{\textbf{Diff}}usion Model for \underline{\textbf{Pose}} Estimation (DiffPose). 
By extending the framework of diffusion model, DiffPose presents a new approach to video-based human pose estimation. Specifically, it reformulates this problem as a conditional generative task of keypoint-wise heatmaps, as illustrated in Fig. \ref{fig:formulation}.
DiffPose consists of two primary stages: a forward diffusion stage that gradually introduces Gaussian noise to the ground truth heatmaps, and a reverse denoising stage that utilizes a Pose-Decoder to recover the original heatmap from the noisy input progressively.

Unlike the vanilla diffusion model \cite{ho2020denoising}, which simply uses U-Net \cite{ronneberger2015u} for denoising, we propose two novel designs that enhance the capabilities of the Pose-Decoder. These modifications enable the Pose-Decoder to better \emph{utilize temporal information} and \emph{focus on joint regions}. 
\textbf{(i)} We design a SpatioTemporal Representation Learner (STRL) which sequentially performs spatial information extraction within each frame and integrates cross-frame knowledge through cascaded Transformers. The resulting features, which contain rich temporal priors, are subsequently utilized as a fixed condition at each denoising step by the Pose-Decoder.
\textbf{(ii)} In addition, we propose a Lookup-based MultiScale Feature Interaction mechanism (LMSFI), which guides the Pose-Decoder to learn intricate representations for pose prediction by inductively leveraging information from both noisy heatmaps and spatiotemporal features.
To be specific, we first construct probabilistic joint fields based on the noisy heatmaps, and  perform lookups  over spatiotemporal features accordingly to activate keypoint region features. 
Then, we model fine-grained correlations between the retrieved local joint features and original global contexts over multiple scales to produce the final representations. 
By conducting feature interaction through LMSFI, we can explicitly reason about the relationships between joints and global contexts. As shown in Figure \ref{fig:feature}, our proposed method is able to learn representations that converge around keypoint regions consistently.


An important feature of the diffusion-based framework is the separation of model training and evaluation. To provide more context, DiffPose is trained to reverse the forward diffusion process (\emph{i.e.}, predict ground truth heatmaps from noises) and performs multi-step denoising to generate predictions based on randomly generated noisy heatmaps at inference. Benefiting from such framework, we demonstrate two distinct properties that appeal to human pose estimation task.
\textbf{(i)} As DiffPose can generate multiple plausible pose estimates by sampling random noises, they can be combined to improve the prediction robustness, especially for challenging joints such as wrists and ankles.
\textbf{(ii)} In contrast to existing methods that adopt a fixed iterative refinement structure \cite{cao2017realtime, wei2016convolutional, newell2016stacked}, DiffPose can adaptively vary the number of denoising steps without retraining the model. 
From extensive experiments, we show that DiffPose consistently outperforms existing well-established approaches on three benchmark datasets. Furthermore, each of our proposed design choices is verified through ablation studies.

The key contributions of this work are summarized as follows: 
(1) To our best knowledge, we are the first to investigate video-based human pose estimation from the lens of generative modeling. In particular, we propose DiffPose, the first model that applies diffusion model to multi-frame human pose estimation.
(2) We demonstrate two properties of DiffPose that are effective on pose estimation: the ability to enhance performance by aggregating multiple pose estimations and to perform flexible iterative refinement without model retraining.
(3) We show that our DiffPose delivers state-of-the-art results on three benchmark datasets, PoseTrack2017, PoseTrack2018, and PoseTrack21.

\section{Related Work}
\textbf{Human pose estimation.}\quad 
Early efforts on human pose estimation focus on static images, starting from building probability graphical structures \cite{kiefel2014human, hara2013computationally} to model the relations between body joints. 
With the advancement of deep learning \cite{he2016deep, vaswani2017attention} and the availability of large-scale benchmark datasets \cite{Iqbal_2017_CVPR, Andriluka_2018_CVPR, doering2022posetrack21}, various deep architectures (\emph{e.g.}, CNN and Transformer)-based methods are currently the dominating solutions \cite{Wei_2016_CVPR,  xiao2018simple, sun2019deep, li2021tokenpose, yuan2021hrformer, yang2021transpose, xu2022vitpose}. There are two mainstream paradigms: i) regressing  the position of keypoints from image directly, and ii) estimating probability heatmaps to represent keypoints locations. 
 Representation using heatmap has gained more popularity due to the performance derived from faster optimization convergence.
	
 Conversely, various studies \cite{song2017thin, pfister2015flowing, liu2021deep, liu2022temporal, wang2020combining} have attempted to estimate human poses in videos.
 \cite{liu2021deep} merges heatmaps of consecutive frames and computes their residuals to obtain joint-level features for pose estimation.
 \cite{liu2022temporal} performs implicit motion compensation using deformable convolutions for better feature aggregation and heatmap prediction. 
 These approaches usually predict a deterministic pose solution for each frame and lack effective recalibrations, which might suffer from localized detection failure especially for challenging joints.
 In contrast, Our approach benefiting from probabilistic diffusion models is able to combine multiple pose solutions naturally to provide more robust estimations. 
 



\textbf{Diffusion model.}\quad
Diffusion models \cite{ho2020denoising, sohl2015deep} are a type of deep generative models that utilize the final state of a Markov chain originating from a standard Gaussian distribution to approximate the distribution of natural images. 
Neural network is typically trained to reverse this diffusion process for each Markov step.
Within this framework, diffusion models have recently demonstrated remarkable results in a wide spectrum of generative tasks from visual images \cite{saharia2022image, avrahami2022blended, gu2022diffusioninst, nichol2021glide, fan2022frido, matsunaga2022fine} to nature language \cite{austin2021structured, gong2022diffuseq, li2022diffusion, huang2022prodiff, popov2021grad, yang2022diffsound}. 
Diffusion models have also proven to be useful in various discriminative computer vision problems \cite{chen2022diffusiondet, amit2021segdiff, baranchuk2021label, gu2022diffusioninst}. 
The pioneer work \cite{amit2021segdiff} presents a diffusion model conditioned on an input image for image segmentation. \cite{chen2022diffusiondet} proposes DiffusionDet which formulates object detection as a generative denoising process from noisy boxes to object boxes. \cite{gu2022diffusioninst} further extends \cite{chen2022diffusiondet} to perform instance segmentation. 
To the best of our knowledge, there have been no previous successful attempts to adapt diffusion models for multi-frame human pose estimation. This paper introduces DiffPose, which explores the potential of diffusion models in video-based human pose estimation and is the first diffusion model to achieve state-of-the-art performance for this task.

\vspace{-1.1em}
\section{Our Approach}

\begin{figure*}
\begin{center}
\includegraphics[width=0.95\linewidth]{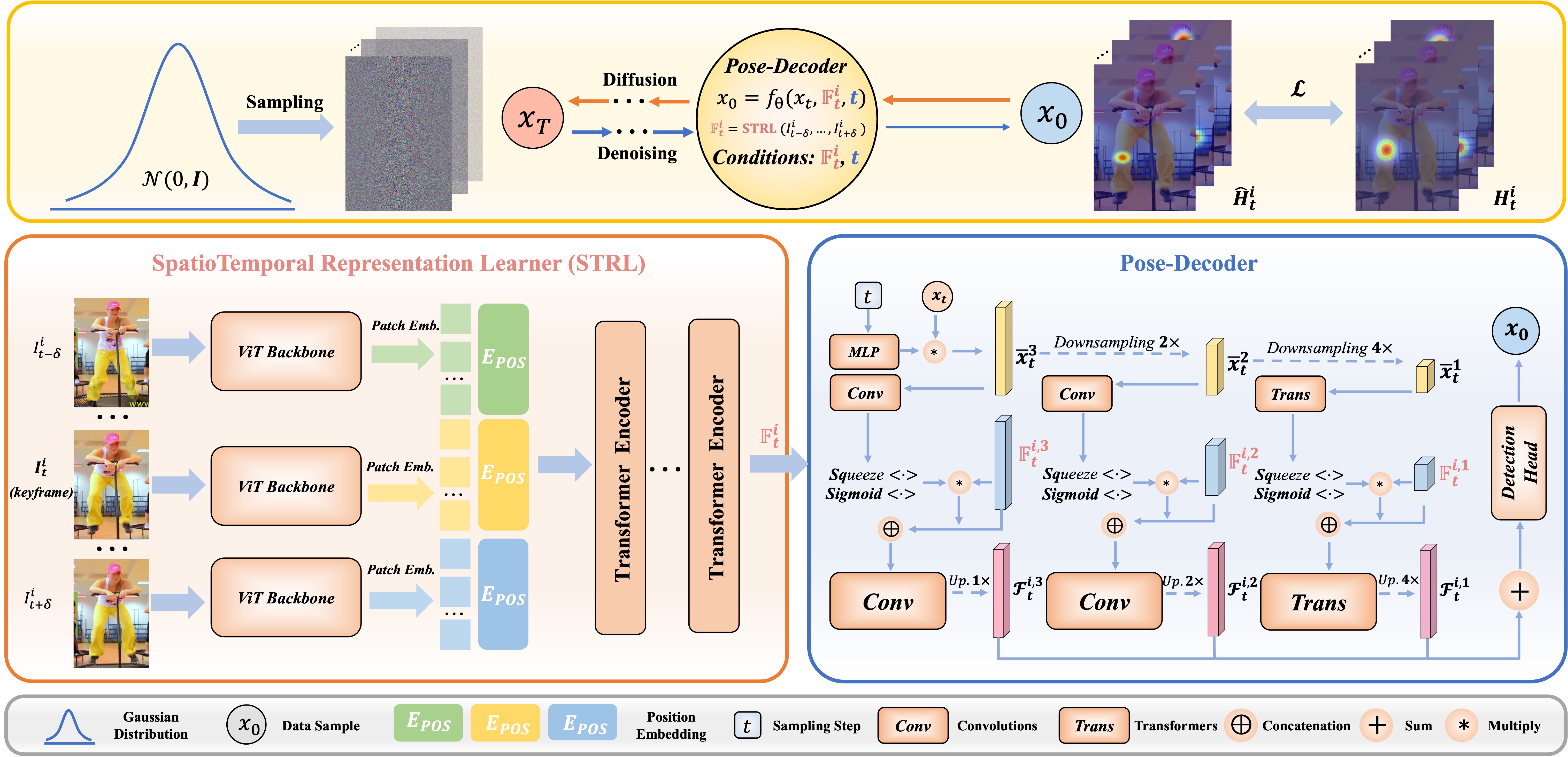}
\end{center}
\vspace{-1em}
\caption{Overall pipeline of the proposed DiffPose framework. The goal is to detect the human pose in the keyframe $I_t^i$. Given an input sequence, our SpatioTemporal Representation Learner (STRL) extracts the spatiotemporal feature $\mathbb{F}_t^i$. The feature $\mathbb{F}_t^i$, the noisy heatmap $\mathbf{x}_t$, and the sampling step $t$ are then passed to the Pose-Decoder, which performs lookup-based multiscale feature interaction to obtain representations $\boldsymbol{\mathcal{F}}_t^{i}=\{{\mathcal{F}}_t^{i,1}, {\mathcal{F}}_t^{i,2}, {\mathcal{F}}_t^{i,3}\}$. Finally, these features are aggregated to attain the final pose estimation $ \hat{\mathbf{H}}_t^i$ (\emph{i.e.}, $\mathbf{x}_0$).}\label{fig:pipeline}
\vspace{-1.2em}
\end{figure*}

\subsection{Preliminaries}
\textbf{Problem Formulation.}\quad 
Following the top-down pose estimation approach, we first obtain all human bounding boxes per frame $I_t$ using an off-the-shelf object detector.
Each bounding box is then enlarged by $25\%$ to crop the same individual in consecutive frames $\boldsymbol{\mathcal{I}_t} = \left \langle I_{t-\delta},...,I_t,...,I_{t+\delta} \right \rangle$ with $\delta$ being a predefined temporal interval. In this way, we obtain the cropped video segment $\boldsymbol{\mathcal{I}_t^i} = \left \langle I_{t-\delta}^i,...,I_t^i,...,I_{t+\delta}^i \right \rangle$ for person $i$. 
Given an image sequence $\boldsymbol{\mathcal{I}_t^i}$ centered on the key frame $I_t^i$, our goal is to estimate the keypoint heatmaps for $I_t^i$.

\textbf{Diffusion Model.}\quad 
Inspired by non-equilibrium thermodynamics \cite{song2019generative, song2020improved}, diffusion models are under the category of latent variable models which aim to reconstruct a task-specific distribution that starts from random noise. 
These models typically consist of two basic processes: 1) a forward process that gradually adds Gaussian noise to sample data, and 2) a reverse process that learns to invert the forward diffusion. 
To be specific, the forward diffusion process is defined as:
\begin{equation}
	\begin{aligned}
		q\left(\mathbf{x}_t |\mathbf{x}_0 \right) := \mathcal{N}\left(\mathbf{x}_t; \sqrt{\bar{\alpha_t}}\mathbf{x}_0, \left(1-\bar{\alpha_t}\right)\mathbf{I} \right), \\
		\mathbf{x}_t = \sqrt{\bar{\alpha_t}}\mathbf{x}_0 + \sqrt{1-\bar{\alpha_t}}\epsilon, \epsilon\sim\mathcal{N}\left(0,1\right),
	\end{aligned}\label{eq1}
\end{equation}
	where $\bar{\alpha_t} := \prod_{s=1}^t \alpha_s = \prod_{s=1}^t (1 - \beta_s)$ and $\beta_s$ denotes the noise variance schedule \cite{ho2020denoising}. The operation in Eq. \ref{eq1} adds noise to the original data sample $\mathbf{x}_0$ and transforms it into a latent noisy sample $\mathbf{x}_t$ at an arbitrary sampling step $t \in \left\{0,1,...,T \right\}$. During training, a neural network $f_\theta(\mathbf{x}_t, t)$ is trained to perform the denoising task either by predicting $\mathbf{x}_0$ or $\epsilon$ (we choose the former as done in \cite{chen2022diffusiondet, chen2022generalist}), with the constraint of ${L}_2$ loss. This process is expressed as:
\begin{equation}
	\begin{aligned}
		\mathcal{L}_{\mathbf{x}_0} = \left\| f_\theta(\mathbf{x}_t, t) - \mathbf{x}_0 \right\|^2.
	\end{aligned}
\end{equation}
	In inference, the learned denoising (reverse) function $f_\theta$ is applied to a random noise sample $\mathbf{x}_T$ along with a preset updating rule \cite{ho2020denoising, song2020denoising}, to reconstruct the data sample $\mathbf{x}_0$ in an iterative way $\mathbf{x}_T \rightarrow \mathbf{x}_{T-\Delta} \rightarrow \cdots \rightarrow \mathbf{x}_0$.

In this paper, we propose a novel framework that enables the diffusion model to better process dynamic contexts for video-based human pose estimation.
Specifically, we present DiffPose which modulates the vanilla diffusion model to \emph{incorporate temporal information} and \emph{attend to keypoint region cues}, resulting in a paradigm more aligned with multi-frame human pose estimation. Our proposed framework is illustrated in Fig. \ref{fig:pipeline}. 
In our problem setting, the original data sample is the ground truth heatmap $\mathbf{x}_0 = \mathbf{H}_t^i$. This heatmap is generated using a 2D Gaussian centered at the annotated joint location. We train a Pose-Decoder $f_\theta(\mathbf{x}_t, \mathbb{F}_t^i, t)$ to recover $\mathbf{x}_0$ from the noisy heatmap $\mathbf{x}_t$ by conditioning on the spatiotemporal feature of the input sequence $\mathbb{F}_t^i$ which is derived by the SpatioTemporal Representation Learner (STRL).

In the following, we first detail the architecture of STRL (Sec. \ref{sec:STRL}) and Pose-Decoder (Sec. \ref{sec:decoder}).
Then, we explain the training and inference algorithms (Sec. \ref{sec:train}) as well as providing discussions on the favorable properties of DiffPose for pose estimation in Sec. \ref{sec:discuss}.

\subsection{Spatiotemporal Representation Learner}\label{sec:STRL}
Inspired by the success of Vision Transformers \cite{dosovitskiy2020image, xu2022vitpose, liu2021swin}, we employ cascaded Transformers to capture the spatial-temporal dependencies among video frames. 
Given the sequence data $\boldsymbol{\mathcal{I}_t^i} = \left \langle I_{t-\delta}^i,...,I_t^i,...,I_{t+\delta}^i \right \rangle$ as input, we first employ a plain Vision Transformer \cite{dosovitskiy2020image, xu2022vitpose} pretrained on ImageNet \cite{deng2009imagenet} as the backbone network to extract spatial features $\left \langle F_{t-\delta}^i,...,F_t^i,...,F_{t+\delta}^i \right \rangle$ for each frame. Subsequently, each frame feature is spatially rearranged and fed into a patch embedding layer, which embeds the feature into tokens $\left \langle \bar{F}_{t-\delta}^i,...,\bar{F}_t^i,...,\bar{F}_{t+\delta}^i \right \rangle$. Then, we concatenate all embedded tokens, retain their spatial information through a learnable position embedding $E_{POS}$, and feed them into cascaded Transformer encoders where each encoder consists of a Multi-Head Self-Attention (MHSA) layer and a feed-forward neural network (FFN). Finally, the encoded deep features of all frames are aggregated via a Multilayer Perceptron (MLP) to produce the spatiotemporal feature $\mathbb{F}_t^i$. The above procedures can be formulated as:
\begin{equation}
	\begin{aligned}	
		\tilde{F}_t^0 &= \mathrm{Concat}\left(\bar{F}_{t-\delta}^i + E_{POS}^{t-\delta},\cdots ,\bar{F}_{t+\delta}^i+ E_{POS}^{t+\delta} \right), \\
		\tilde{F}_t^{'l} &= \tilde{F}_t^{l-1} + \mathrm{MHSA}(\mathrm{LN}(\tilde{F}_t^{l-1})),\\
		\tilde{F}_t^{l} &= \tilde{F}_t^{'l} + \mathrm{FFN}(\mathrm{LN}(\tilde{F}_t^{'l})),\\
		& \qquad\vdots\\
		 \mathbb{F}_t^i &= \mathrm{MLP}(\mathrm{LN}(\tilde{F}_t^{L})),
	\end{aligned}
\end{equation}
where the superscript $l \in [1,2,...,L]$ denotes the output of $l$-$th$ Transformer layer and $\tilde{F}_t^0$ represents the initial feature. The function $\rm LN(\cdot)$ indicates the LayerNorm layer. Note that the spatial (\emph{i.e.}, the number of tokens) and channel dimensions within each transformer layer remain constant. 

\subsection{Pose-Decoder}\label{sec:decoder}
After obtaining the spatiotemporal feature $\mathbb{F}_t^i$, the Pose-Decoder denoises the heatmap $\mathbf{x}_t$ by taking $\mathbb{F}_t^i$ together with the sampling step $t$ as conditions, and output the predicted heatmap $\hat{\mathbf{x}}_0=\hat{\mathbf{H}}_t^i$. Specifically, we first project the step index $t$ into an embedding and utilize the embedding to rescale the initial noisy heatmap $\mathbf{x}_t$, attaining the step-adaptive version $\bar{\mathbf{x}}_t$. Then, we model the global correlations between $\mathbb{F}_t^i$ and $\bar{\mathbf{x}}_t$ across multiple scales via Transformer or convolutional structures, and obtain multi-scale representations $\boldsymbol{\mathcal{F}}_t^{i}=\{{\mathcal{F}}_t^{i,1}, {\mathcal{F}}_t^{i,2}, {\mathcal{F}}_t^{i,3}\}$. Finally, these features are integrated and passed to a detection head to predict the pose heatmap $\hat{\mathbf{H}}_t^i$.

In order to encourage the representations $\boldsymbol{\mathcal{F}}_t^{i}$ to focus on keypoint regions, we propose a Lookup-based MultiScale Feature Interaction mechanism (LMSFI) to inductively model correlations between $\mathbb{F}_t^i$ and $\bar{\mathbf{x}}_t$. It consists of two procedures: pairwise size-matched feature generation and lookup-based feature interaction.

\textbf{Pairwise size-matched feature generation.}\quad Given $\mathbb{F}_t^i \in \mathbb{R}^{C\times H \times W}$ and $\bar{\mathbf{x}}_t \in \mathbb{R}^{c\times 4H \times 4W}$ with different spatial dimensions, we  perform upsampling and downsampling separately to construct size-matched feature pairs $\langle\mathbb{F}_t^i,\bar{\mathbf{x}}_t\rangle$. Specifically, we adopt several deconvolution layers to perform $1\times$, $2\times$, and $4\times$ upsampling of resolution over $\mathbb{F}_t^i$, and obtain corresponding features $\{\mathbb{F}_t^{i,1}, \mathbb{F}_t^{i,2}, \mathbb{F}_t^{i,3}\}$. Similarly, stride convolutions are used to downsample on $\bar{\mathbf{x}}_t$ to produce $\{\bar{\mathbf{x}}^1_t, \bar{\mathbf{x}}^2_t, \bar{\mathbf{x}}^3_t\}$ ($\bar{\mathbf{x}}_t = \bar{\mathbf{x}}^3_t$). With the above process, we attain multi-scale size-matched feature pairs $\langle\mathbb{F}_t^{i,J}, \bar{\mathbf{x}}_t^J\rangle$. The superscript $J=\{1,2,3\}$ refers to resolutions of different levels from low to high.

\textbf{Lookup-based feature interaction.}\quad Upon constructing multi-scale feature pairs $\langle\mathbb{F}_t^{i,J}, \bar{\mathbf{x}}_t^J\rangle$,  we model interactions between the spatiotemporal feature $\mathbb{F}_t^{i,j}$ and the noisy heatmap $\bar{\mathbf{x}}_t^j$ at each resolution $j$ individually to obtain corresponding feature representation $\mathcal{F}_t^{i,j}$. A naive approach would be to \emph{directly} concatenate and aggregate $\mathbb{F}_t^{i,j}$ and $\bar{\mathbf{x}}_t^j$. In our experiments, we show that the learned keypoint features of this scheme are scattered across significant areas (see Fig. \ref{fig:feature}), resulting in performance reduction as shown in Table \ref{abl-components}.
In practice, heatmaps reveal the likelihood of the locations containing joints, whereas the noisy heatmap $\bar{\mathbf{x}}_t^j$ is corrupted and can provide a negligible amount of valid real-valued information \cite{chen2022analog}. As a result, directly modeling correlations of $\bar{\mathbf{x}}_t^j$ and $\mathbb{F}_t^{i,j}$ is extremely challenging. Therefore, we adopt an inductive modeling strategy which first performs lookups over the spatiotemporal feature $\mathbb{F}_t^{i,j}$ according to the heatmap  $\bar{\mathbf{x}}_t^j$ to retrieve local joint feature $\bar{\mathbb{F}}_t^{i,j}$, and then models correlations between the local feature $\bar{\mathbb{F}}_t^{i,j}$ and the vanilla global context $\mathbb{F}_t^{i,j}$.

More specifically, considering that the computational complexity of self-attention increases quadratically with input resolution, we adopt a composite structure that uses Transformers and convolutions to capture feature interactions at low and high resolutions, respectively. \textbf{(i)} For the low resolution feature pair $\langle\mathbb{F}_t^{i,1}, \bar{\mathbf{x}}_t^1\rangle$, the noisy heatmap $\bar{\mathbf{x}}_t^1$ is first embedded to feature tokens, and a Transformer encoder is leveraged to perform self-refinement to yield $\bar{\bar{\mathbf{x}}}_t^1$. Then, we take the maximum activations along the depth dimension over $\bar{\bar{\mathbf{x}}}_t^1$ to squeeze the global channel information into a single-channel descriptor, followed by a sigmoid function to obtain an attention mask $A^1$ that indicates possible keypoint fields. Then, the mask $A^1$ is used to retrieve corresponding spatiotemporal features $\bar{\mathbb{F}}_t^{i,1}$. Finally, $\bar{\mathbb{F}}_t^{i,1}$ and $\mathbb{F}_t^{i,1}$ are concatenated and processed by cascaded Transformers, followed by upsampling of resolution to output feature ${\mathcal{F}}_t^{i,1}$. The above process can be described as:
\begin{equation}
	\begin{aligned}
		\bar{\bar{\mathbf{x}}}_t^1 &= \mathrm{SeRef}(\bar{\mathbf{x}}_t^1), & A^1 &= \mathrm{Sigmoid} (\mathrm{Sq}(\bar{\bar{\mathbf{x}}}_t^1 )), \\
		\bar{\mathbb{F}}_t^{i,1} &= A^1 \odot \mathbb{F}_t^{i,1}, &  \mathcal{F}_t^{i,1} &= \mathrm{Up}(\mathrm{Trans}(\bar{\mathbb{F}}_t^{i,1} \oplus \mathbb{F}_t^{i,1})),
	\end{aligned}
\end{equation}
where $\mathrm{SeRef}(\cdot)$, $\mathrm{Sq}(\cdot)$, $\odot$,  $\oplus$, and $\mathrm{Up}(\cdot)$ denote the operations of self-refinement, squeezing, spatial-wise multiplication, concatenation, and upsampling, respectively.  \textbf{(ii)} For high-resolution feature pairs $\langle\mathbb{F}_t^{i,j}, \bar{\mathbf{x}}_t^j\rangle$ with $j=2,3$, an  analogical procedure is executed using convolutions: 
\begin{equation}
	\begin{aligned}
		\bar{\bar{\mathbf{x}}}_t^j &= \mathrm{Conv}(\bar{\mathbf{x}}_t^j), & A^j &= \mathrm{Sigmoid} (\mathrm{Sq}(\bar{\bar{\mathbf{x}}}_t^j )), \\
		\bar{\mathbb{F}}_t^{i,j} &= A^j \odot \mathbb{F}_t^{i,j}, &  \mathcal{F}_t^{i,j} &= \mathrm{Up}(\mathrm{Conv}(\bar{\mathbb{F}}_t^{i,j} \oplus \mathbb{F}_t^{i,j})).
	\end{aligned}
\end{equation}

\textbf{Heatmap generation.}\quad Ultimately, we integrate feature representations across all scales  $\{{\mathcal{F}}_t^{i,1}, {\mathcal{F}}_t^{i,2}, {\mathcal{F}}_t^{i,3}\}$ via element-wise addition, and employ a detection head (\emph{i.e.}, a $3\times 3$ convolution) to yield the predicted heatmap $\hat{\mathbf{H}}_t^i$. By inductively modeling multi-scale feature interactions, our Pose-Decoder is able to reason about the fine-grained relations of keypoints and global contexts, thereby producing more tailored representations that attend to joint areas.

\subsection{Overall Training and Inference Algorithms}\label{sec:train}
\textbf{Training.}\quad We perform diffusion process that corrupts ground truth heatmaps to noisy heatmaps, and train the Pose-Decoder for  heatmap denoising to reverse this process. The overall training procedure of our DiffPose is provided in Algorithm \ref{alg:training} in the Appendix. Specifically, we sample Gaussian noises according to $\alpha_t$ in Eq. \ref{eq1} and add them to ground truth heatmaps to obtain the noisy samples. The parameter $\alpha_t$ at each sampling step $t$ is predefined by a monotonically decreasing cosine scheme, as adopted in \cite{ho2020denoising}. We employ the standard pose estimation loss (mean square error) to supervise the model training:
\begin{equation}
	\begin{aligned}
		\mathcal{L} = \left\|\mathbf{H}_t^i - \hat{\mathbf{H}}_t^i \right\|_2^2,
	\end{aligned}
\end{equation}
where $\mathbf{H}_t^i$ and $\hat{\mathbf{H}}_t^i$ denote the ground truth and predicted pose heatmaps, respectively.

\textbf{Inference.}\quad
The proposed DiffPose conducts denoising on noisy heatmaps sampled from Gaussian distribution, progressively refining its predictions over multiple sampling steps. For each sampling step, the Pose-Decoder takes random noisy heatmaps or the predicted heatmaps of the last sampling step as input and outputs the estimated heatmaps of the current step. Then, we adopt DDIM \cite{song2020denoising} to update the heatmaps for the next step. Detailed inference procedure is provided in Algorithm \ref{alg:Inference}.

\subsection{Discussion}\label{sec:discuss}
Building on diffusion-based architecture, our proposed framework DiffPose is able to decouple training and testing stages which enables a more adaptable inference process. By extending this concept, we investigate further and demonstrate the unique benefits of DiffPose for pose estimation, specifically in the areas of \emph{flexible pose ensemble} and \emph{flexible iterative refinement}.

\textbf{Flexible pose ensemble.} In common practices, one usually performs inference starting with a single initial sample. However, the diffusion model is intrinsically probabilistic \cite{amit2021segdiff} and can generate diverse outputs for different noise inputs. Correspondingly, taking different noisy heatmaps as input, the DiffPose can yield different plausible pose predictions that possess respective keypoint recognition preferences. Ensembling these complementary pose solutions can enhance the robustness and stability of model predictions especially for challenging joints. To exploit this phenomenon, we initialize $N$ groups of noisy heatmaps for inference and subsequently average their predictions. Experimental results show that the complementary pose ensemble brings significant performance improvement (see Table \ref{abl-n}).

\textbf{Flexible iterative refinement.} After training the model, the DiffPose performs multi-step refinement (sampling) progressively to yield the final pose prediction. In practice, the number of sampling steps can be adjusted flexibly without retraining the model, which is preferable to the prior approaches that adopt a fixed structure of iterative refinement \cite{cao2017realtime, wei2016convolutional, luo2018lstm, newell2016stacked}. By increasing iterative sampling steps, the resulting representations would be more delicate, which fosters accurate pose estimation (Fig. \ref{fig:feature}).

\renewcommand\arraystretch{1.1}
\begin{table}
  \resizebox{0.48\textwidth}{!}{
  \begin{tabular}{l|ccccccc|c}
    \hline
      Method                            &Head   &Shoulder &Elbow       &Wrist   &Hip    &Knee   &Ankle   &{\bf Mean}\cr
      \hline
      PoseTracker \cite{girdhar2018detect}   &$67.5$ &$70.2$   &$62.0$      &$51.7$  &$60.7$ &$58.7$ &$49.8$  &{$60.6$}\cr
     PoseFlow \cite{xiu2018pose}         &$66.7$ & $73.3$  &$68.3$      &$61.1$  &$67.5$ &$67.0$ &$61.3$  &{$ 66.5$}\cr
JointFlow \cite{doering2018joint}        & -     & -       &-           &-       &-      &-      &-       &{ $ 69.3$}\cr
   FastPose \cite{zhang2019fastpose}   	&$80.0$ &$80.3$   &$69.5$      &$59.1$  &$71.4$ &$67.5$ &$59.4$  &{$ 70.3$}\cr
   TML++ \cite{hwang2019pose}    	 		&-       &-     &-      &-      &-      &-       &-    &{$ 71.5$}\cr
Simple (R-50) \cite{xiao2018simple}    &$79.1$ &$80.5$   &$75.5$      &$66.0$  &$70.8$ &$70.0$ &$61.7$  &{$72.4$}\cr
Simple (R-152) \cite{xiao2018simple}    &$81.7$ &$83.4$   &$80.0$      &$72.4$  &$75.3$ &$74.8$ &$67.1$  &{$ 76.7$}\cr
  STEmbedding \cite{jin2019multi}        &$83.8$ &$81.6$   &$77.1$      &$70.0$  &$77.4$ &$74.5$ &$70.8$  &{$ 77.0$}\cr
        HRNet \cite{sun2019deep}         &$82.1$ &$83.6$   &$80.4$      &$73.3$  &$75.5$ &$75.3$ &$68.5$  &{$ 77.3$}\cr
         MDPN \cite{guo2018multi}        &$85.2$ &$88.5$   &$83.9$      &$77.5$  & $79.0$&$77.0$ &$71.4$  &{$ 80.7$}\cr
   CorrTrack \cite{rafi2020self}   &$86.1$ &$87.0$   &$83.4$      &$76.4$  & $77.3$&$79.2$ &$73.3$  &{$ 80.8$}\cr 
   Dynamic-GNN \cite{yang2021learning} 	 &$88.4$ &$88.4$   &$82.0$      &$ 74.5$ &$79.1$ &$78.3$ &$73.1$  &{ $81.1$}\cr
   PoseWarper \cite{bertasius2019learning} &$81.4$ &$88.3$   &$83.9$      &$ 78.0$ &$82.4$ &$80.5$ &$73.6$  &{ $ 81.2$}\cr
   DCPose \cite{liu2021deep}  &$ 88.0$  &$ 88.7$     &$ 84.1$   &$78.4$&$ 83.0$        &$ 81.4$&$ 74.2$ &$ 82.8$\cr
   DetTrack \cite{wang2020combining}  &$89.4$       &$89.7$     &$85.5$ &$79.5$ &$82.4$      &$80.8$       &$76.4$   &$83.8$\cr
    FAMI-Pose \cite{liu2022temporal}	&$\bf 89.6$  &$ 90.1$ &$ 86.3$ &$80.0$ &$ 84.6$ &$83.4$ &$ 77.0$ &$ 84.8$\cr
    \hline 
     \rowcolor{gray!20} \bf DiffPose (Ours)	&$ 89.0$  &$\bf 91.2$ &$\bf 87.4$ &$\bf 83.5$ &$\bf 85.5$ &$\bf 87.2$ &$\bf 80.2$ &$\bf 86.4$\cr 
    \hline
    \end{tabular}}
    \vspace{.1em}
    \caption{Quantitative results on the \textbf{PoseTrack2017} validation set.} \label{17val}
\end{table}

\renewcommand\arraystretch{1.1}
\begin{table}
   \resizebox{0.48\textwidth}{!}{
   \begin{tabular}{l|ccccccc|c}
     \hline
      Method                            &Head &Shoulder &Elbow  &Wrist &Hip &Knee &Ankle &{\bf Mean}\cr
     \hline
  STAF \cite{raaj2019efficient}    	  	&-       &-     &-      &$64.7$ &-      &-       &$62.0$   &{$70.4$}\cr
 AlphaPose \cite{fang2017rmpe}           &$63.9$  &$78.7$&$77.4$ &$71.0$ &$73.7$ &$73.0$    &$69.7$     &{$71.9$}\cr
  TML++ \cite{hwang2019pose}    	 		&-       &-     &-      &-      &-      &-       &-    &{$ 74.6$}\cr
 MDPN \cite{guo2018multi}                &$75.4$ &$81.2$ &$79.0$ &$74.1$ &$72.4$ &$73.0$  &$69.9$   &{$75.0$}\cr
 PGPT \cite{bao2020pose}    	 		&-       &-     &-      &$72.3$ &-      &-       &$72.2$   &{$76.8$}\cr
 Dynamic-GNN \cite{yang2021learning} 	&$80.6$ &$84.5$   &$80.6$  &$ 74.4$ &$75.0$ &$76.7$ &$71.8$  &{ $77.9$}\cr
 PoseWarper \cite{bertasius2019learning} &$79.9$&$86.3$&$82.4$&$77.5$&$79.8$&$78.8$&$73.2$  &{ $79.7$}\cr
 PT-CPN++ \cite{yu2018multi} &$82.4$ &$88.8$ &$86.2$ &$79.4$ &$72.0$ &$80.6$ &$76.2$  &$80.9$\cr
 DCPose \cite{liu2021deep}&$ 84.0$ &$ 86.6$&$ 82.7$&$ 78.0$&$ 80.4$&$ 79.3$&$ 73.8$&$ 80.9$\cr 
 DetTrack \cite{wang2020combining}  &$84.9$ &$87.4$ &$84.8$ &$79.2$ &$77.6$      &$79.7$       &$75.3$   &$81.5$ \cr
 FAMI-Pose \cite{liu2022temporal} &$\bf 85.5$&$\bf 87.7$&$ 84.2$&$ 79.2$&$ \bf 81.4$&$81.1$&$ 74.9$&$ 82.2$\cr
	 \hline

    \rowcolor{gray!20}  \bf DiffPose (Ours)&$ 85.0$&$\bf 87.7$&$\bf 84.3$&$\bf 81.5$&$\bf 81.4$&$\bf 82.9$&$\bf 77.6$&$\bf 83.0$\cr
     \hline
     \end{tabular}}
     \vspace{.1em}
     \caption{Quantitative results on the \textbf{PoseTrack2018} validation set.}
     \vspace{-.2em}\label{18val}
   \end{table}

\renewcommand\arraystretch{1.2}
\begin{table}
   \resizebox{0.48\textwidth}{!}{
   \begin{tabular}{l|ccccccc|c}
     \hline
      Method                            &Head &Shoulder &Elbow  &Wrist &Hip &Knee &Ankle &{\bf Mean}\cr
     \hline
   Tracktor++  w. poses \cite{bergmann2019tracking, doering2022posetrack21}   	  	&-       &-     &-      &- &-      &-       &-   &{$71.4$}\cr
  CorrTrack \cite{rafi2020self, doering2022posetrack21}    	  	&-       &-     &-      &- &-      &-       &-   &{$72.3$}\cr
  CorrTrack w. ReID \cite{rafi2020self, doering2022posetrack21}    	  	&-       &-     &-      &- &-      &-       &-   &{$72.7$}\cr
  Tracktor++ w. corr. \cite{bergmann2019tracking, doering2022posetrack21}    	  	&-       &-     &-      &- &-      &-       &-   &{$73.6$}\cr
 DCPose \cite{liu2021deep}&$ 83.2$&$ 84.7$&$ 82.3$&$ 78.1$&$ 80.3$&$ 79.2$&$ 73.5$&$ 80.5$\cr 
 FAMI-Pose \cite{liu2022temporal} &$ 83.3$&$ 85.4$&$ 82.9$&$ 78.6$&$ 81.3$&$80.5$&$ 75.3$&$ 81.2$\cr
	 \hline

     \rowcolor{gray!20} \bf DiffPose (Ours)&$\bf 84.7$&$\bf 85.6$&$\bf 83.6$&$\bf 80.8$&$\bf 81.4$&$\bf 83.5$&$\bf 80.0$&$\bf 82.9$\cr
     \hline
     \end{tabular}}
     \vspace{.1em}
      \caption{Quantitative results on the \textbf{PoseTrack21} dataset.} \vspace{-1em}\label{21val}
   \end{table}

\section{Experiments}
\subsection{Experimental Settings}
\textbf{Datasets.}\quad 
We benchmark the proposed DiffPose on three widely used benchmark datasets for video-based human pose estimation, PoseTrack2017 \cite{Iqbal_2017_CVPR}, PoseTrack2018 \cite{Andriluka_2018_CVPR}, and PoseTrack21 \cite{doering2022posetrack21}. These datasets contain video sequences of tricky scenarios where clustered people perform rapid movement. Specifically, \textbf{PoseTrack2017} includes $250$ video clips for training and $50$ videos for validation (split according to the official protocol), with a total of $80,144$ pose annotations. \textbf{PoseTrack2018} considerably increases the number of clips, containing $593$ videos for training, $170$ videos for validation, and a total of $153, 615$ pose annotations.  Both datasets identify $15$ keypoints, with an additional label for joint visibility. The training videos are densely annotated in the center $30$ frames, and validation videos are additionally labeled every four frames. \textbf{PoseTrack21} further enriches and refines PoseTrack2018 especially for annotations of small persons and persons in crowds, including  $177,164$ human pose annotations.  

\textbf{Evaluation metric.}\quad
The performance of the proposed method is evaluated with the widely-adopted \cite{wang2020combining, sun2019deep, bertasius2019learning, liu2021deep} human pose estimation metric namely average precision (\textbf{AP}). We compute the AP for each joint and then average overall joints to obtain the final performance (\textbf{mAP}).  

\textbf{Implementation details.}\quad Our DiffPose is implemented with PyTorch. The input image size is fixed to $256 \times 192$. We incorporate data augmentation including random rotation $[-45^{\circ}, 45^{\circ}]$, random scale $[0.65, 1.35]$,  truncation (half body), and flipping during training. The time interval $\delta$ is set to $2$. We define the total sampling steps $T=1000$. We adopt the AdamW \cite{reddi2019convergence} optimizer with a base learning rate of $5e-4$ (decays to $5e-5$ and $5e-6$ at the $20^{th}$ and $40^{th}$epochs, respectively). We train the model using $4$ TITAN RTX GPUs. All training process is terminated within $60$ epochs. During inference, we initialize $N=10$ groups of noises, and set the iterative denoising steps to $4$.

\begin{figure*}
\begin{center}
\includegraphics[width=0.94\linewidth]{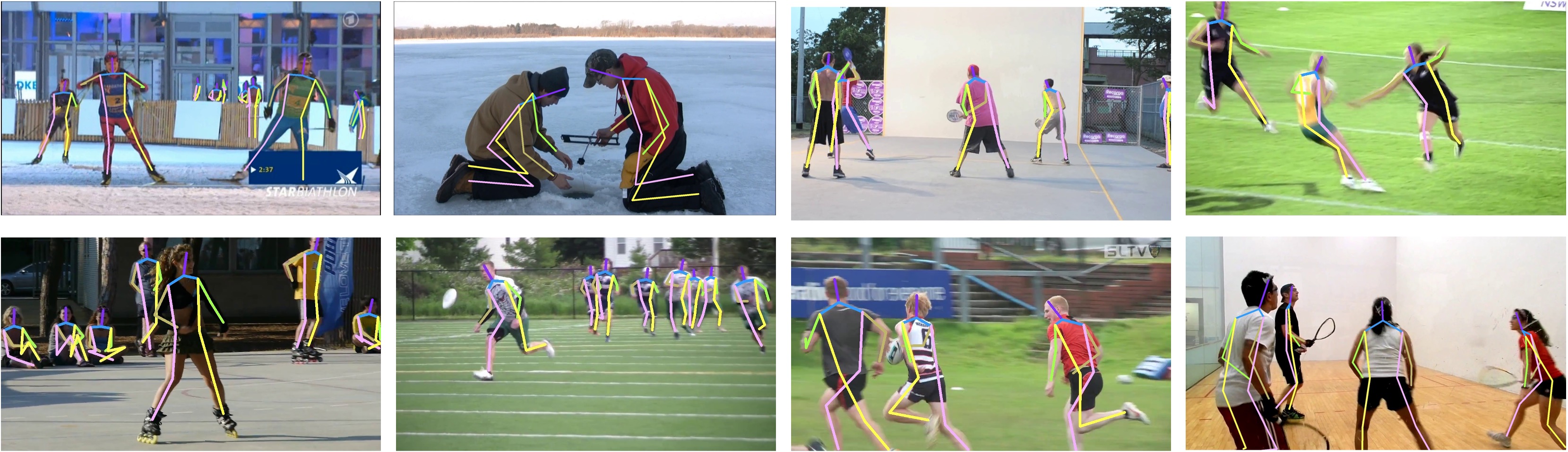}
\end{center}
\vspace{-1em}
\caption{Visual results of our DiffPose on benchmarks. Challenging scenarios including fast motion and mutual occlusion are involved.}
 \vspace{-1.em}
\label{fig:results}
\end{figure*}

\subsection{Comparison with State-of-the-art Approaches}
\textbf{Results on the PoseTrack2017 dataset.}\quad
We first benchmark our model on the PoseTrack2017 dataset. A total of $17$ methods are compared, and their performances on the validation set are summarized in Table \ref{17val}. 
We can observe that our DiffPose delivers state-of-the-art pose estimation performance compared to existing well-established approaches, by adopting a  generative paradigm for the first time.
DiffPose attains the final performance of $86.4$ mAP, and provides a $1.6$ mAP gain over the previous best-performed method FAMI-Pose \cite{liu2022temporal}. The performance boost for challenging joints (\emph{i.e.}, wrist, ankle) is also encouraging: we achieve an mAP of $83.5$ ($\uparrow 3.5$) for wrists and an mAP of $80.2$ ($\uparrow 3.2$) for ankles. Such consistent performance improvements suggest the great potential of diffusion models in pose estimation. Another observation is that pose estimation methods that integrate temporal information (such as DetTrack and FAMI-Pose) indeed surpass approaches using only the single keyframe. This corroborates the importance of our design that injects spatiotemporal features into the DiffPose model. Furthermore, we show example visualization results in Fig. \ref{fig:results}, which are indicative of the robustness of our method in tricky scenes.

\textbf{Results on the PoseTrack2018 dataset.}\quad
We further evaluate the proposed DiffPose on the PoseTrack2018 dataset, and report the detailed results on the validation set in Table \ref{18val}. As shown in this table,  our DiffPose once again outperforms all other approaches and delivers the best results. We obtain the final performance of $83.0$ mAP, with an mAP of $84.3$, $81.5$, $82.9$, and $77.6$ for the elbow, wrist, knee, and ankle joints.

\textbf{Results on the PoseTrack21 dataset.}\quad
Performance comparisons of our model and previous state-of-the-art methods on the PoseTrack21 dataset are provided in Table \ref{21val}. We observe that existing method FAMI-Pose \cite{liu2022temporal} has already achieved an impressive performance of $81.2$ mAP. In contrast, our DiffPose is able to achieve $82.9$ ($\uparrow 1.7$) mAP. We also obtain an mAP of $80.8$ for the wrist joint and $80.0$ for the ankle joint. 
   
 \renewcommand\arraystretch{1.0}
\begin{table}
   \resizebox{0.48\textwidth}{!}{
   \begin{tabular}{c|cc|c}
     \hline
         Method &Spatiotemporal. (STRL)  &Lookup. (LMSFI) &Mean\cr
    \hline
    Direct-single & &   & $52.5$\cr
    (a) & &\checkmark  & $82.4$\cr
    (b) &\checkmark &\checkmark  & $\bf 86.4$\cr
	 \hline
     \end{tabular}}
     \vspace{.1em}
      \caption{Ablation of different components in DiffPose.} \label{abl-components}\vspace{-.5em}
   \end{table}
 
\renewcommand\arraystretch{1.0}
\begin{table}
   \resizebox{0.48\textwidth}{!}{
   \begin{tabular}{c|cc|c}
     \hline
         Method &Multi-scale feature  &Aggregation &Mean\cr
    \hline
    Low-resolution & &   & $84.4$\cr
    (a) &\checkmark &\checkmark  & $\bf 86.4$\cr
    (b) &\checkmark &\checkmark $\star$  & $ 86.2$\cr
	 \hline
     \end{tabular}}
     \vspace{.1em}
      \caption{Ablation of various designs in LMSFI. $\star$ denotes fusing multi-scale features with concatenation and aggregation.} \label{abl-LMSFI}\vspace{-1.5em}
   \end{table}

\begin{table*}
	\begin{minipage}[t]{0.45\linewidth}
	\vspace{0pt}
    \centering 
	\resizebox{1\textwidth}{!}{
    \begin{tabular}{c|ccccccc|c}
    \hline
      Method &Head   &Shoulder &Elbow       &Wrist   &Hip    &Knee   &Ankle   &{\bf Mean}\cr
      \hline
    FAMI-Pose \cite{liu2022temporal}	&$ 89.6$  &$ 90.1$ &$ 86.3$ &$80.0$ &$ 84.6$ &$83.4$ &$ 77.0$ &$ 84.8$\cr
    \hline 
	\bf $N=1$	&$ 85.4$  &$ 87.0$ &$ 80.1$ &$ 74.6$ &$ 78.1$ &$ 82.6$ &$74.1$ &$ 80.4$\cr 
	\bf $N=5$	&$ 87.4$  &$ 88.8$ &$ 83.7$ &$ 79.6$ &$ 83.7$ &$84.8$ &$ 78.1$ &$ 84.0$\cr 
	 \bf $N=10$, DiffPose	&$ 89.0$  &$ 91.2$ &$ 87.4$ &$ 83.5$ &$ 85.5$ &$ 87.2$ &$ 80.2$ &$ \bf 86.4$\cr 
    \hline
    \end{tabular}}
    \vspace{.1em}
    \caption{Ablation of modifying the number of initial noises $N$.} \vspace{-1.2em}\label{abl-n}
  \end{minipage}
   \begin{minipage}[t]{0.55\linewidth} 
 	\vspace{0pt}
	 \centering
 	\resizebox{1\textwidth}{!}{
 	 \begin{tabular}{c|ccccccc|c}
    \hline
      Sampling Steps &Head   &Shoulder &Elbow       &Wrist   &Hip    &Knee   &Ankle   &{\bf Mean}\cr
       \hline 
	\bf $Steps=1$	&$ 88.9$  &$ 90.4$ &$ 85.9$ &$ 82.5$ &$ 85.1$ &$ 86.6$ &$80.2$ &$ 85.9$\cr 
	\bf $Steps=2$	&$ 89.0$  &$ 91.0$ &$ 87.0$ &$ 83.4$ &$ 85.4$ &$87.1$ &$ 80.2$ &$ 86.3$\cr 
	 \bf $Steps=4$, DiffPose	&$ 89.0$  &$ 91.2$ &$ 87.4$ &$ 83.5$ &$ 85.5$ &$ 87.2$ &$ 80.2$ &$\bf 86.4$\cr 
    \hline
    \end{tabular}}
    \vspace{.1em}
    \caption{Ablation of modifying the number of denoising steps.} \vspace{-1.2em}\label{abl-steps}
 \end{minipage}
\end{table*}

\subsection{Ablation Study}
We perform ablation experiments to investigate the contribution of each component in our DiffPose, including SpatioTemporal Representation Learner (STRL) as well as the Lookup-based MultiScale Feature Interaction mechanism (LMSFI). We also examine the efficacy of various design choices in LMSFI. Finally, we study the influence of modifying the number of initial noise heatmaps and iterative steps (during inference) on the final performance.

\textbf{Study on components of DiffPose.}\quad
We empirically evaluate the effectiveness of each proposed component, and report the results in Table \ref{abl-components}. We first construct a simple baseline namely Direct-single, which takes the single keyframe as condition and directly concatenates and aggregates noisy heatmap with image features to model their interactions. This straightforward scheme produces a severely degraded pose estimation performance of $52.5$ mAP. This is in line with our intuitions, \emph{i.e.}, the noisy heatmap is corrupted and usually contains distracting information, which incurs inherent difficulties in directly learning the correlations between noisy heatmaps and image features. \textbf{(a)} For this setting, we introduce the proposed LMSFI into the baseline Direct-single. Remarkably, the LMSFI inductively models feature interactions and improves performance from an mAP of $52.5$ to $82.4$. This significant performance boost ($\uparrow 29.9$ mAP) corroborates the importance of our LMSFI in guiding the model to focus on specific joint regions. \textbf{(b)} The final setting further incorporates spatiotemporal cues as the sampling condition and corresponds to our full DiffPose. The performance improvement of $4.0$ mAP demonstrates the effectiveness of our DiffPose in introducing temporal information to facilitate video-based pose estimation.


\textbf{Study on Lookup-based MultiScale Feature Interaction.}\quad
We further explore the influence of various designs within LMSFI, and tabulate the results in Table \ref{abl-LMSFI}. We  model feature interactions in a single low resolution to form the baseline Low-resolution. \textbf{(a)} We introduce multi-scale  fusion to the baseline method (\emph{i.e.}, the complete DiffPose) and produce the final performance of $86.4$ ($\uparrow 2.0$) mAP. This significant performance improvement upon incorporation of the multi-scale fusion highlights its effectiveness in learning informative representations for better performance. \textbf{(b)} We also examine the impact of fusing multi-scale features by concatenation and aggregation. The result in mAP ($86.2$) changes marginally.

\textbf{Study on initial noises.}\quad
As discussed in Sec. \ref{sec:discuss}, we propose a pose ensemble strategy that randomly initializes $N$ groups of Gaussian noises during inference and averages their predictions. Table \ref{abl-n} shows the effects of adopting different $N$, where $N$ is set to $1$, $5$, and $10$. The quantitative results in mAP reflect a gradual performance improvement with increasing initial noises, from $80.4 \rightarrow 84.0 \rightarrow 86.4$. This phenomenon can be  attributed to the probabilistic nature of the diffusion model, whereby DiffPose is able to forecast diverse plausible poses from different  noises. Ensembling such pose solutions enhances the robustness of model predictions and significantly boosts the pose estimation performance. Another observation is that the improvement in mAP with increasing $N$ mainly stems from challenging joints such as wrists ($\uparrow 8.9$ mAP) and ankles ($\uparrow 6.1$ mAP), and this fact still remains valid when compared to  FAMI-Pose \cite{liu2022temporal}. This suggests that the proposed pose ensemble strategy derived from the diffusion-based architecture can potentially facilitate the pose detection in intractable scenes (\emph{e.g.}, occlusions, blur).

\textbf{Study on denoising steps.}\quad
As discussed in Sec. \ref{sec:discuss}, DiffPose can adopt an arbitrary number of iterative sampling steps. To investigate how the number of iterative steps affects the final performance, we experiment with $Steps \in \{1, 2, 4\}$ and report the results in Table \ref{abl-steps}. It is clear that more iteration steps result in better performances. This is in accordance with our expectations, \emph{i.e.}, the captured features are progressively refined to focus on keypoint regions upon multiple iterations, leading to better results.

\begin{figure}
\begin{center}
\includegraphics[width=0.98\linewidth]{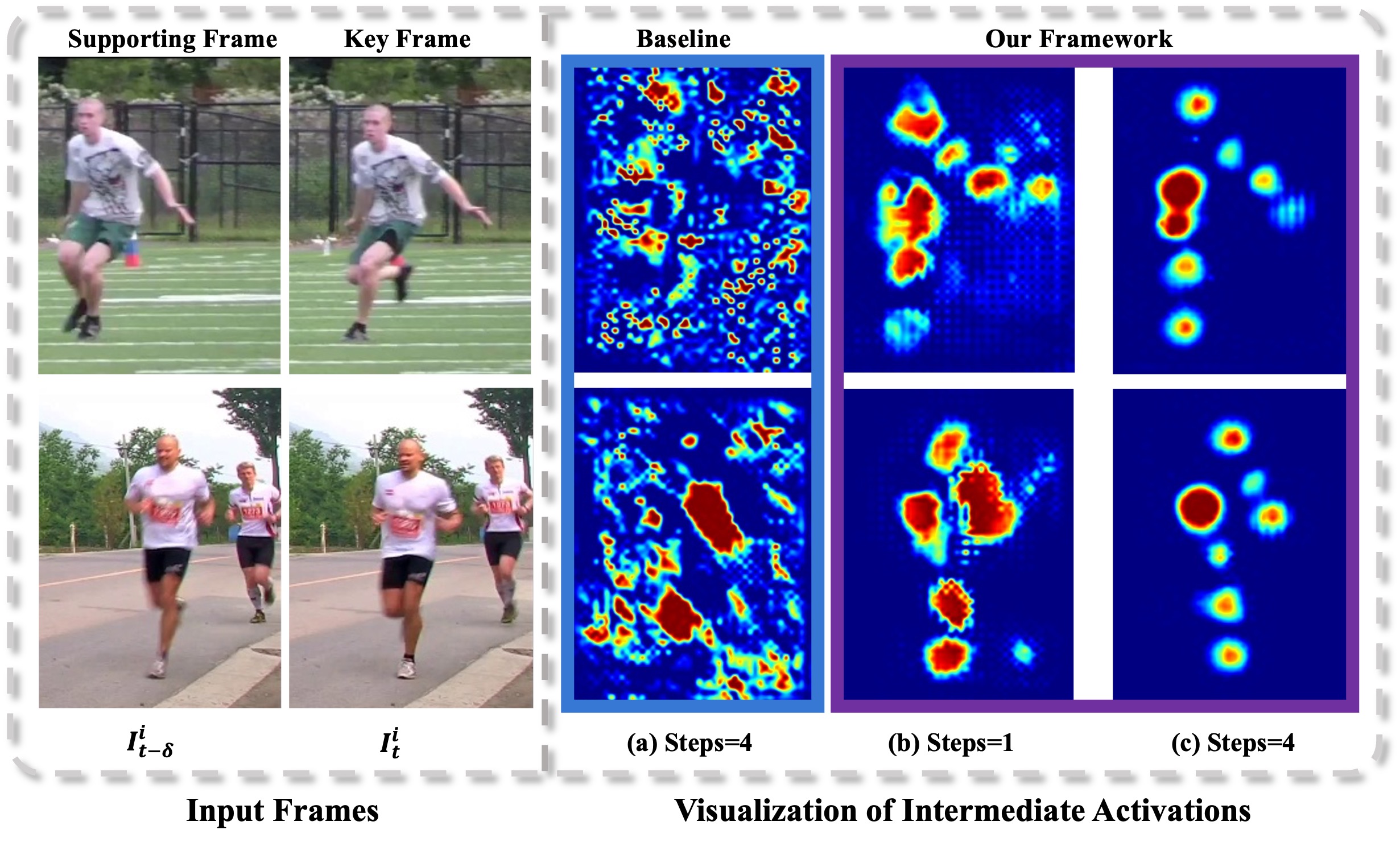}
\end{center}
\vspace{-1em}
\caption{Visualizations of intermediate activations of the Direct-multi baseline (a) and our DiffPose at different denoising steps (b) $Steps=1$ and (c) $Steps=4$.}
\label{fig:feature}
\vspace{-1em}
\end{figure}

\subsection{Qualitative Analyses on DiffPose}\label{qua}

\textbf{Representation visualization.}\quad In addition to the quantitative results, we also provide qualitative analyses to better understand the mechanism behind DiffPose. Fig. \ref{fig:feature} displays the intermediate activations of the Direct-multi baseline (incorporating STRL into the Direct-single) as well as our DiffPose. We observe that the DiffPose produces compact representations (b) and (c) that attend to local keypoint regions, while the features derived from the baseline (a) spread across salient areas. This provides empirical evidence that our LMSFI is effective in learning tailored representations for pose estimation. On the other hand, the features upon multi-step refinement are more attentive that encompass less task-irrelevant information.

\textbf{Visual comparisons.}\quad
We further examine the ability of our model in dealing with challenging scenarios such as mutual occlusion and fast motion. We depict in Fig. \ref{fig:qualitative} the side-by-side comparisons of a) our DiffPose against state-of-the-art methods b) FAMI-Pose \cite{liu2022temporal} and c) HRNet \cite{sun2019deep}. It is observed that our DiffPose consistently produces more robust pose predictions for various challenging scenes. HRNet \cite{sun2019deep} is designed for static images and dose not incorporate temporal cues, leading to suboptimal results in degraded frames. On the other hand, FAMI-Pose \cite{liu2022temporal} adopts a deterministic pose estimation paradigm, yielding a single pose solution for each person. Through the principled design of the model architecture (\emph{i.e.}, STRL, LMSFI) as well as the flexible training and testing pipeline of the diffusion model, our DiffPose is more adept at handling tricky cases.

\begin{figure}
\begin{center}
\includegraphics[width=0.98\linewidth]{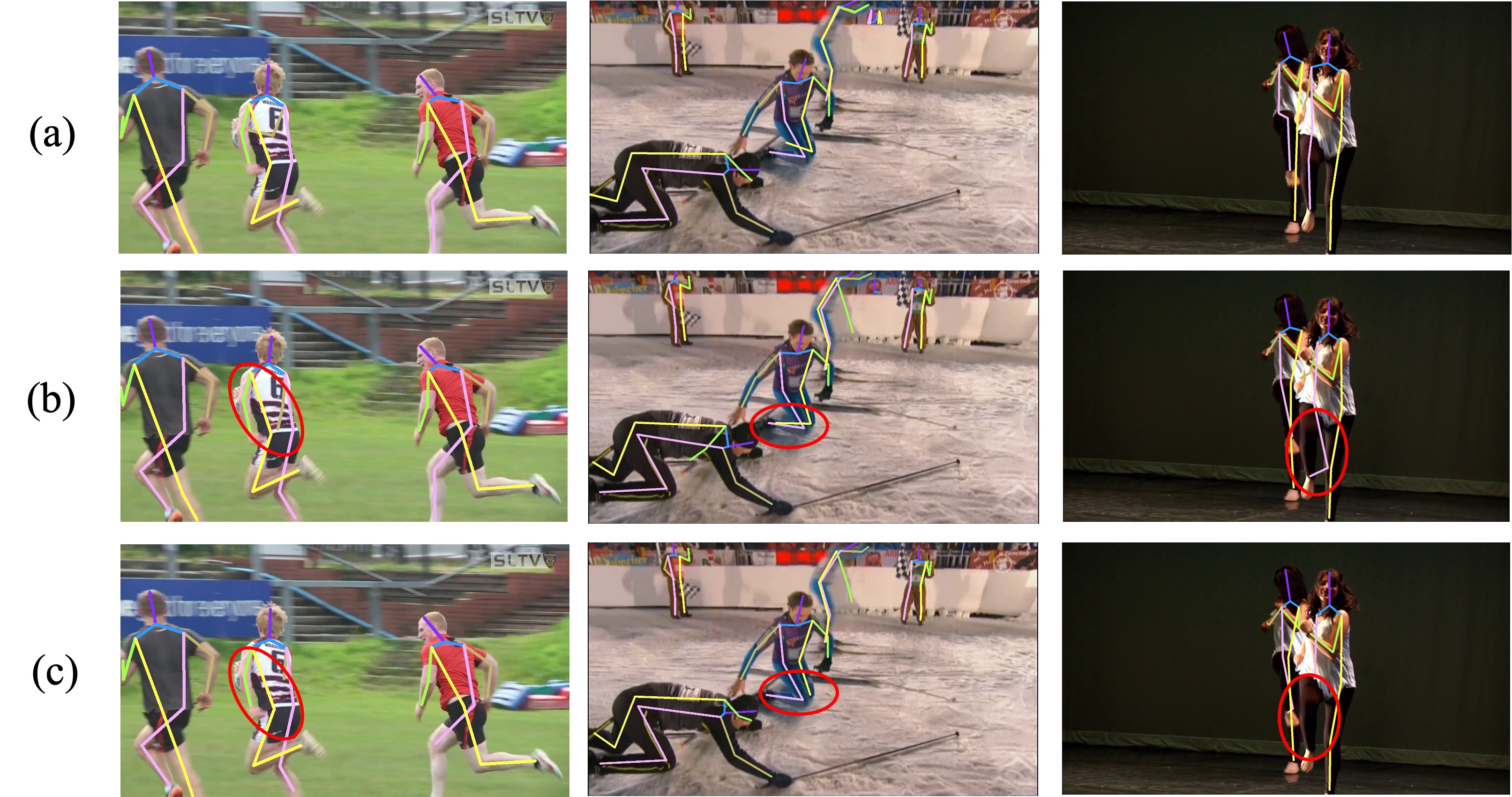}
\end{center}
\vspace{-.5em}
\caption{Visual comparisons of the pose estimation results of our DiffPose (a), FAMI-Pose (b),  and HRNet-W48 (c) on the challenging cases from PoseTrack dataset. Inaccurate detections are highlighted by the red circles.}
\label{fig:qualitative}
\vspace{-1.1em}
\end{figure}

\section{Conclusion and Future Works}
In this paper, we explore the video-based human pose estimation task from the perspective of generative modeling. We present a novel framework termed DiffPose which treats multi-frame human pose estimation as a conditional generative process of keypoint heatmaps. We design a SpatioTemporal Representation Learner (STRL) to integrate temporal clues into the diffusion model, as well as a Lookup-based MultiScale Feature Interaction mechanism (LMSFI) for inducing the model to attend to keypoint regions. Furthermore, we show two attractive properties of DiffPose for pose estimation including flexible pose ensemble and iterative refinement, which enable enhanced performance without retraining the model. Empirical evaluations on three standard benchmark datasets, PoseTrack2017, PoseTrack2018, and PoseTrack21 demonstrate that DiffPose achieves state-of-the-art performance. Future works include applying DiffPose to other vision tasks such as 3D human pose estimation and pose tracking, and refining the pipeline for accelerated inference. 

\section{Acknowledgements}
This work is supported in part by the National Natural Science Foundation of China under grant No. 62203184 and the International Cooperation Project under grant No. 20220402009GH.
This work is also supported in part by the MSIT (Ministry of Science and ICT), Korea, under the ITRC (Information Technology Research Center) support program (IITP-2023-2020-0-01789), supervised by the IITP (Institute for Information \& Communications Technology Planning \& Evaluation).

{\small
\bibliographystyle{ieee_fullname}
\bibliography{References}

\clearpage
\section*{Appendix}
\section*{Training and Inference Algorithms}
In this section, we present detailed training and inference algorithms of the proposed DiffPose framework.

\textbf{Training.}\quad 
In the training phase, we perform the diffusion process that corrupts ground truth heatmaps $\mathbf{x_0}$ to noisy heatmaps $\mathbf{x_t}$ , and train the Pose-Decoder $f_\theta(\cdot)$ to reverse this process. Algorithm \ref{alg:training} provides the overall training procedure.

\textbf{Inference.}\quad
Algorithm \ref{alg:Inference} summarizes the detailed inference procedure of the proposed DiffPose, which can be regarded as an iterative heatmap refinement process. The parameter $steps$ denotes the number of iterative denoising steps. Specifically, for each sampling step, the Pose-Decoder takes random noisy heatmaps or the predicted heatmaps of the last sampling step as input and outputs the estimated heatmaps of the current step. Then, we adopt DDIM to update the heatmaps for the next step.

\begin{algorithm}[!h]
	\renewcommand{\algorithmicrequire}{\textbf{Input:}}
	\caption{DiffPose Training}
	\label{alg:training}
	\begin{algorithmic}[1]
		\REQUIRE $\text{image\_sequence}:\boldsymbol{\mathcal{I}_t^i}, \text{gt\_heatmap}:\mathbf{H_t^i}$
		\REPEAT
		\STATE { $\boldsymbol{\mathbb{F}_t^i} = \textbf{STRL}(\boldsymbol{\mathcal{I}_t^i})$ }
		\STATE { $\mathbf{x}_0 = \mathbf{H_t^i}$}
		\STATE { $t \sim \text{Uniform}(\{1, ... ,T\})$ }
		\STATE { $\epsilon \sim \mathcal{N}\left(0,1\right)$}
		\STATE { $\mathbf{x}_t = \sqrt{\bar{\alpha_t}}\mathbf{x}_0 + \sqrt{1-\bar{\alpha_t}}\epsilon$ }
		\STATE { Take gradient descent step on \\
		\qquad $\Delta\theta\left\|f_\theta(\mathbf{x}_t, \boldsymbol{\mathbb{F}_t^i}, t)-\mathbf{x}_0 \right\|^2$
		}
		\UNTIL{converged}		
	\end{algorithmic}  
\end{algorithm}

\begin{algorithm}[!h]
	\renewcommand{\algorithmicrequire}{\textbf{Input:}}
	\renewcommand{\algorithmicensure}{\textbf{Output:}}
	\caption{DiffPose Inference}
	\label{alg:Inference}
	\begin{algorithmic}[1]
		\REQUIRE $\text{image\_sequence}:\boldsymbol{\mathcal{I}_t^i}, steps, T$
	    \ENSURE $\text{predicted\_heatmap}:\mathbf{\hat{H}^i_t}$

		\STATE { $\boldsymbol{\mathbb{F}_t^i} = \textbf{STRL}(\boldsymbol{\mathcal{I}_t^i})$ }
		\STATE { $\mathbf{x}_t\sim \mathcal{N}\left(0,1\right)$}
		\STATE { $\text{times} = \text{Reversed}(\text{Linespace}(-1,T,steps))$}
		\STATE { $\text{time\_pairs} = \text{List}(\text{Zip}(\text{times}[:-1],\text{times}[1:]))$}
		\FOR {$t_{now}$,$t_{next}$ \textbf{in} \text{time\_pairs}}
		\STATE {$\mathbf{\hat{H}^i_t} = f_\theta(\mathbf{x}_t, \boldsymbol{\mathbb{F}_t^i}, t_{now})$}
		\STATE {$\mathbf{x}_t = \textbf{DDIM}(\mathbf{x}_t, \mathbf{\hat{H}^i_t}, t_{now}, t_{next})$}
		\ENDFOR
		\RETURN $\mathbf{\hat{H}^i_t}$
	\end{algorithmic}  
\end{algorithm}

\section*{Supplementary Experiments}
In this section, we investigate the influence of the temporal interval $\delta$ within the SpatioTemporal Representation Learner (STRL). We further display more visualized results of the DiffPose model.

\renewcommand\arraystretch{0.98}
\begin{table}[h]
   \resizebox{0.48\textwidth}{!}{
   \begin{tabular}{c|cc|c}
     \hline
         Method &STRL, $\delta=0$  &STRL, $\delta=1$ &STRL, $\delta=2$\cr
    \hline
    Mean &$82.4$  & $84.5$  & $\bf 86.4$\cr
	 \hline
     \end{tabular}}
     \vspace{.1em}
      \caption{Ablation of modifying the temporal interval $\delta$.} \label{abl-STRL}
   \end{table}

\textbf{Study on temporal interval $\delta$.}
We examine the effects of adopting different temporal interval $\delta$ that controls the number of supporting frames within the STRL. From the results in Table \ref{abl-STRL}, we observe a performance improvement with higher number of supporting frames, whereby the mAP increases from $82.4$ for $\delta=0$ to $84.5$, $86.4$ at $\delta=1$,  $\delta=2$, respectively. This is in accordance with our expectation, \emph{i.e.}, incorporating more frames brings up abundant temporal contexts which foster accurate pose estimation. Note that we can further increase $\delta$ to obtain better performance. In view of the trade-off between performance and computation load, we choose $\delta=2$.

\begin{figure*}[!ht]
\begin{center}
\includegraphics[width=0.98\linewidth]{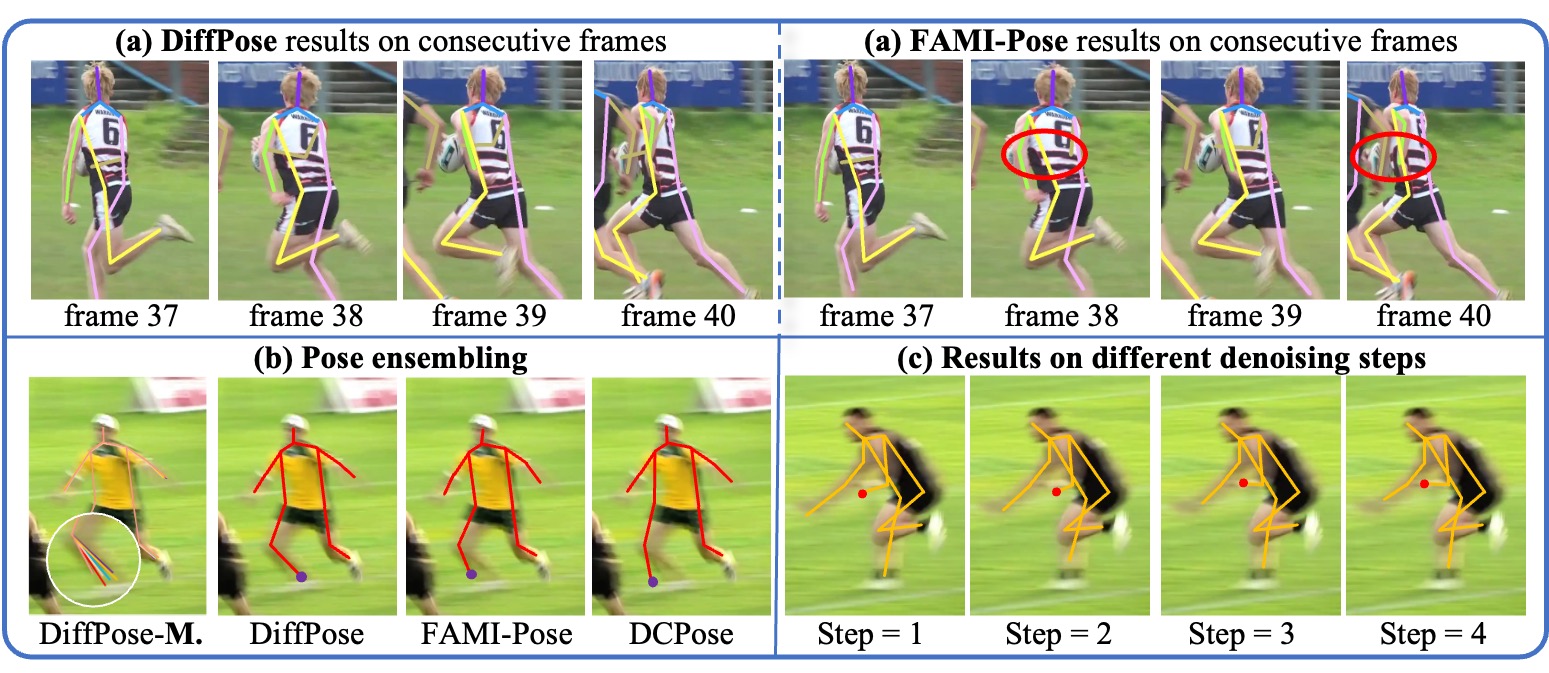}
\end{center}
\vspace{-1.em}
\caption{Visualization examples. ``DiffPose-M." denotes the multiple pose predictions in pose ensembling. Best view in color.
}
\vspace{-1em}
\label{fig:vis}
\end{figure*}

\textbf{Visualized results.}\quad  Video comparisons and visualization of pose ensembling and gradual denoising are depicted in Fig. \ref{fig:vis} (a), (b), and (c), respectively. Inaccurate estimations are highlighted in (a). Furthermore, we present more visualized results of the proposed DiffPose.  Figs. \ref{fig:17}--\ref{fig:21} capture the pose estimation results on PoseTrack2017, PoseTrack2018, and PoseTrack21 datasets, respectively. We clearly observe that our DiffPose achieves accurate and robust pose estimations in various challenging scenes.

\begin{figure*}
\begin{center}
\includegraphics[width=0.94\linewidth]{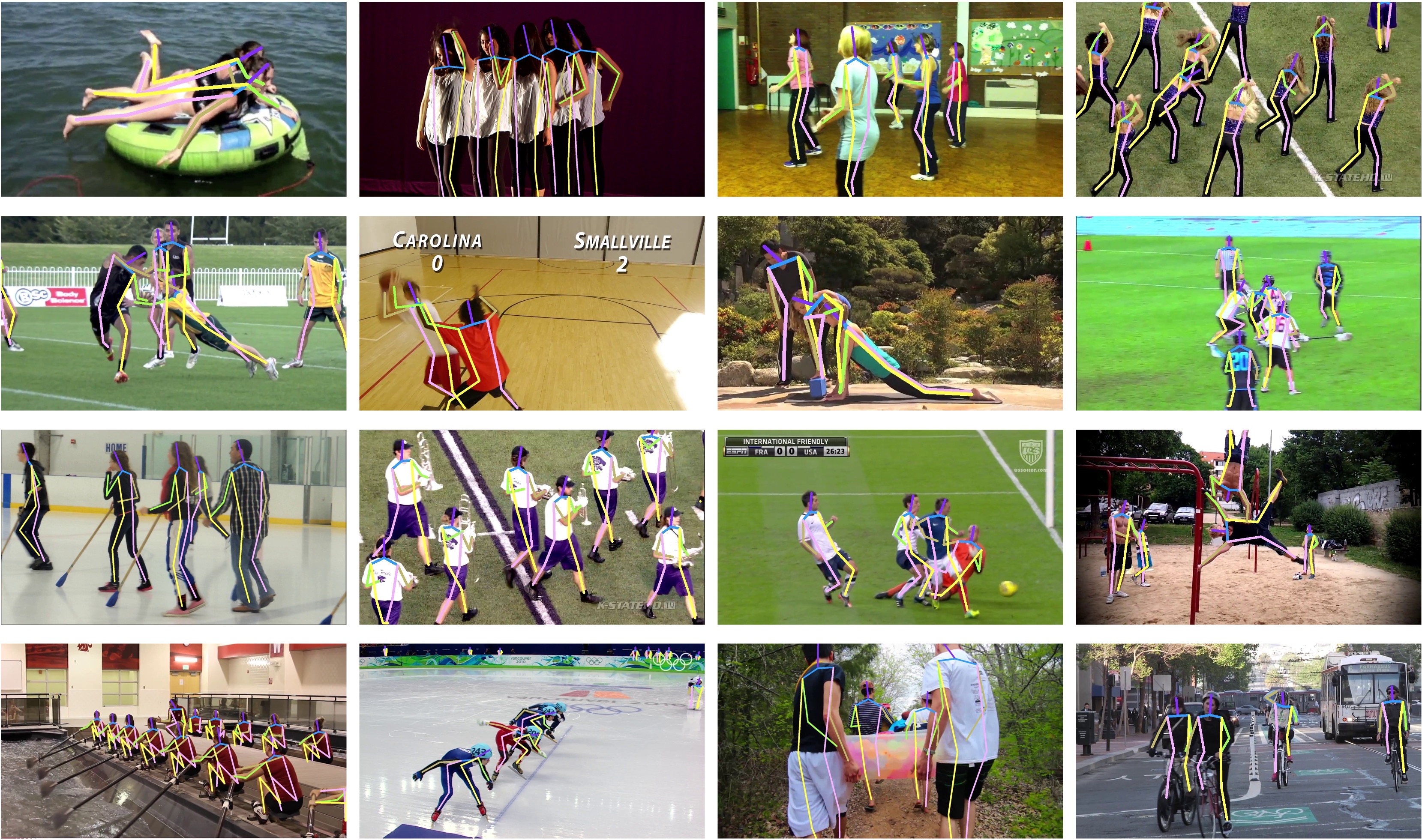}
\end{center}
\vspace{-1em}
\caption{Visual results of our DiffPose on PoseTrack2017. Challenging scenes such as fast motion or occlusions are involved.}
\label{fig:17}
\end{figure*}

\begin{figure*}
\begin{center}
\includegraphics[width=0.94\linewidth]{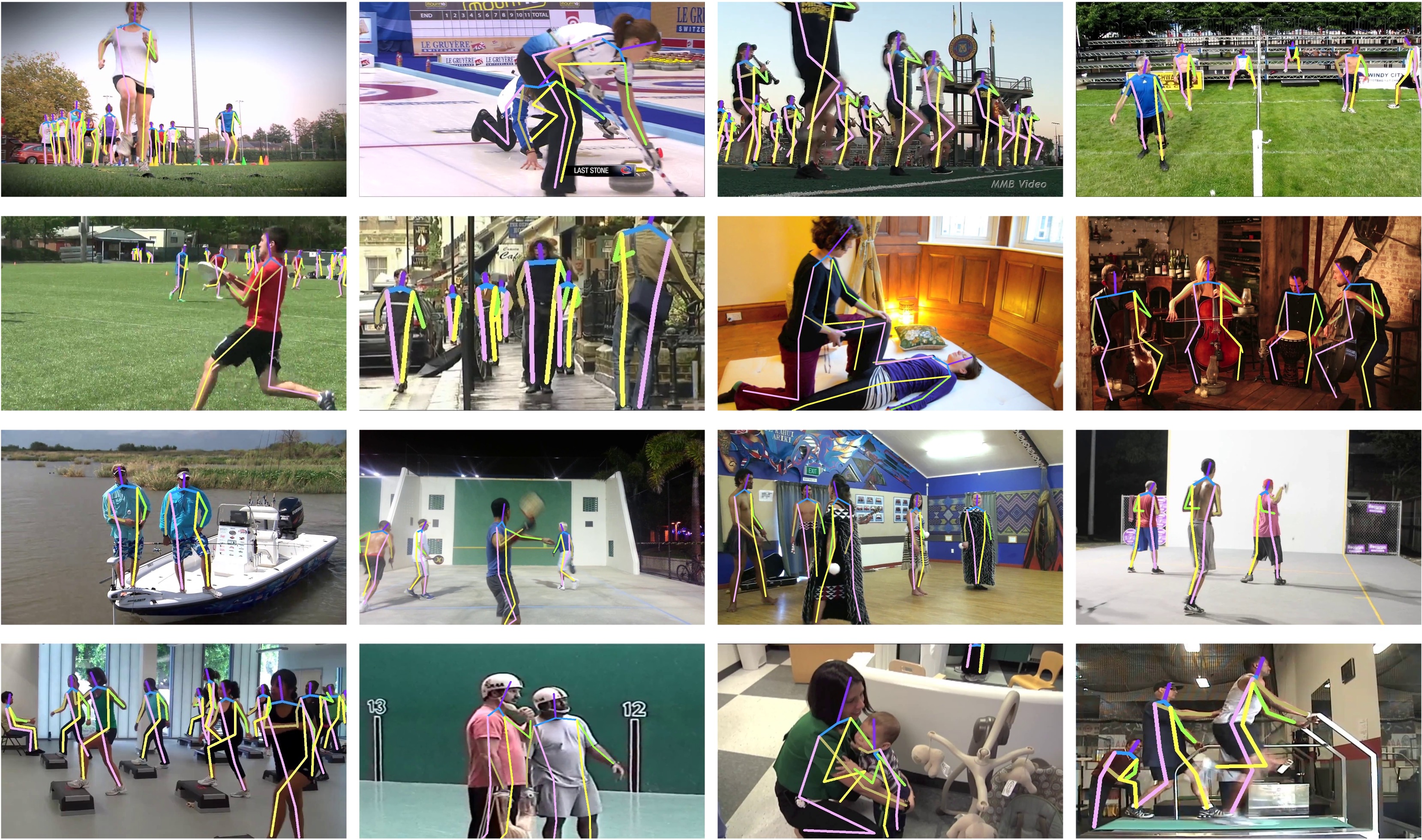}
\end{center}
\vspace{-1em}
\caption{Visual results of our DiffPose on PoseTrack2018. Challenging scenes such as fast motion or occlusions are involved.}
\label{fig:18}
\end{figure*}

\begin{figure*}
\begin{center}
\includegraphics[width=0.94\linewidth]{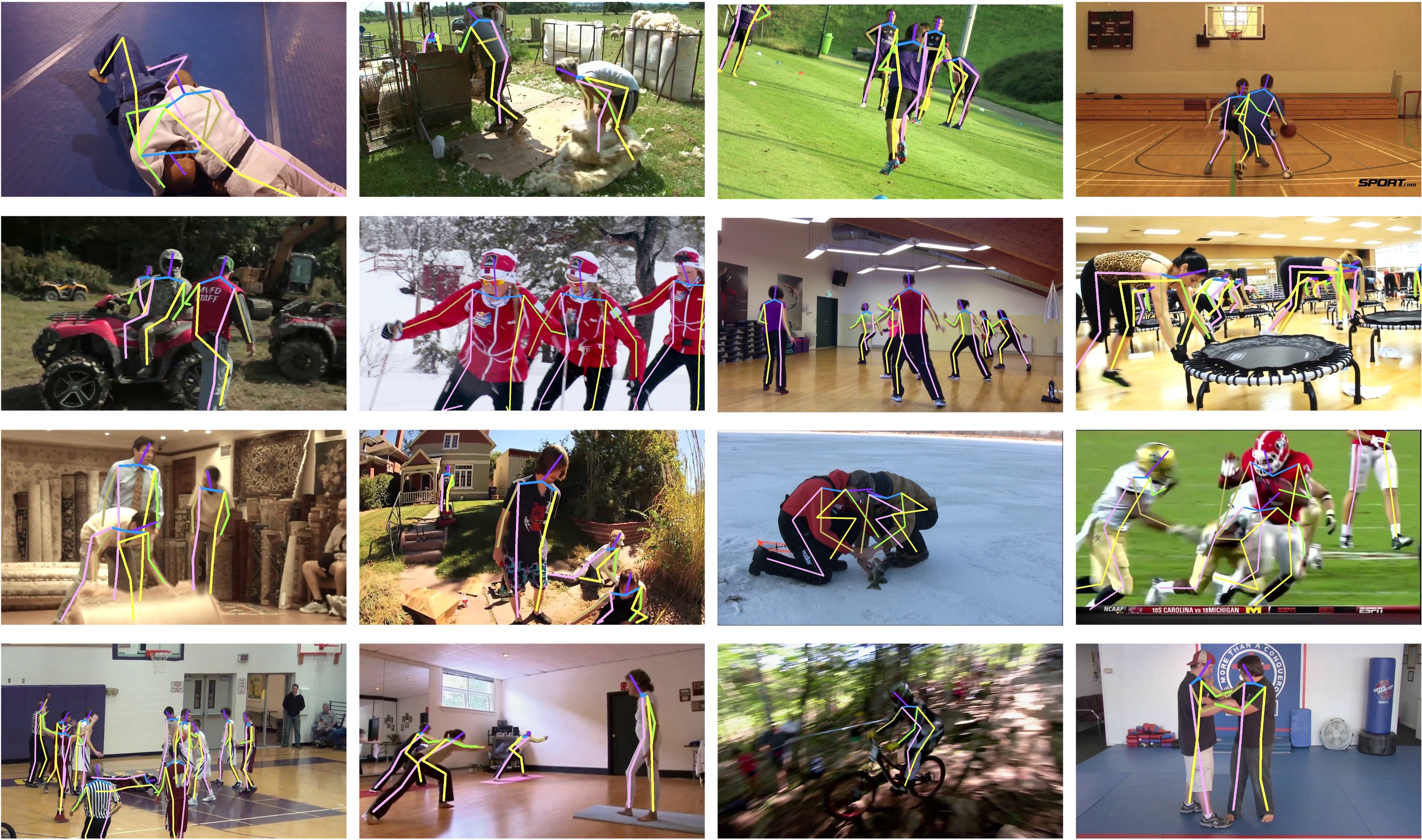}
\end{center}
\vspace{-1em}
\caption{Visual results of our DiffPose on PoseTrack21. Challenging scenes such as fast motion or occlusions are involved.}
\label{fig:21}
\end{figure*}

\end{document}